\documentclass[11pt]{article}

\usepackage[margin=1in]{geometry}  
\usepackage{graphicx}             
\usepackage[numbers,sort&compress]{natbib} 
\usepackage{mhchem}               
\usepackage{amsmath}              
\usepackage{booktabs}             
\usepackage[utf8]{inputenc}       
\usepackage[T1]{fontenc}        
\usepackage{hyperref}             
\hypersetup{
    colorlinks=true,
    linkcolor=blue,
    filecolor=magenta,      
    urlcolor=blue,
    citecolor=blue,
}
\usepackage{authblk}              
\usepackage{epstopdf}             

\title{\textbf{Large language models in materials science and the need for open-source approaches}}

\author[1]{Fengxu Yang}
\author[2]{Weitong Chen}
\author[1*]{Jack D. Evans}

\affil[1]{\textit{School of Physics, Chemistry and Earth Sciences, The University of Adelaide, Adelaide 5005, Australia}}
\affil[2]{\textit{School of Computer and Mathematical Sciences, The University of Adelaide, Adelaide 5005, Australia}}
\affil[*]{\textit{Corresponding author: j.evans@adelaide.edu.au}}

\date{} 

\begin{document}

\maketitle

\begin{abstract}
Large language models (LLMs) are rapidly transforming materials science.
This review examines recent LLM applications across the materials discovery pipeline, focusing on three key areas: mining scientific literature , predictive modelling, and multi-agent experimental systems.
We highlight how LLMs extract valuable information such as synthesis conditions from text, learn structure-property relationships, and can coordinate agentic systems integrating computational tools and laboratory automation.
While progress has been largely dependent on closed-source commercial models, our benchmark results demonstrate that open-source alternatives can match performance while offering greater transparency, reproducibility, cost-effectiveness, and data privacy.
As open-source models continue to improve, we advocate their broader adoption to build accessible, flexible, and community-driven AI platforms for scientific discovery.
\end{abstract}


\section{Introduction}
Large Language Models (LLMs) have been evolving at an unprecedented pace over the last few years and have shown remarkable capabilities across a wide range of tasks and domains.\cite{ai4science_impact_2023, zimmermann_34_2025} Trained on immense and diverse data from across different knowledge areas, they have developed an extraordinary ability to understand, process, and generate complex text.\cite{grigorov_what_2024} Beyond general-purpose applications such as education\cite{kasneci_chatgpt_2023} and health\cite{thirunavukarasu_large_2023}, LLMs are increasingly emerging as powerful tools for 
scientific research.

\begin{figure}[htbp] 
    \centering
    \includegraphics[width=0.4\textwidth]{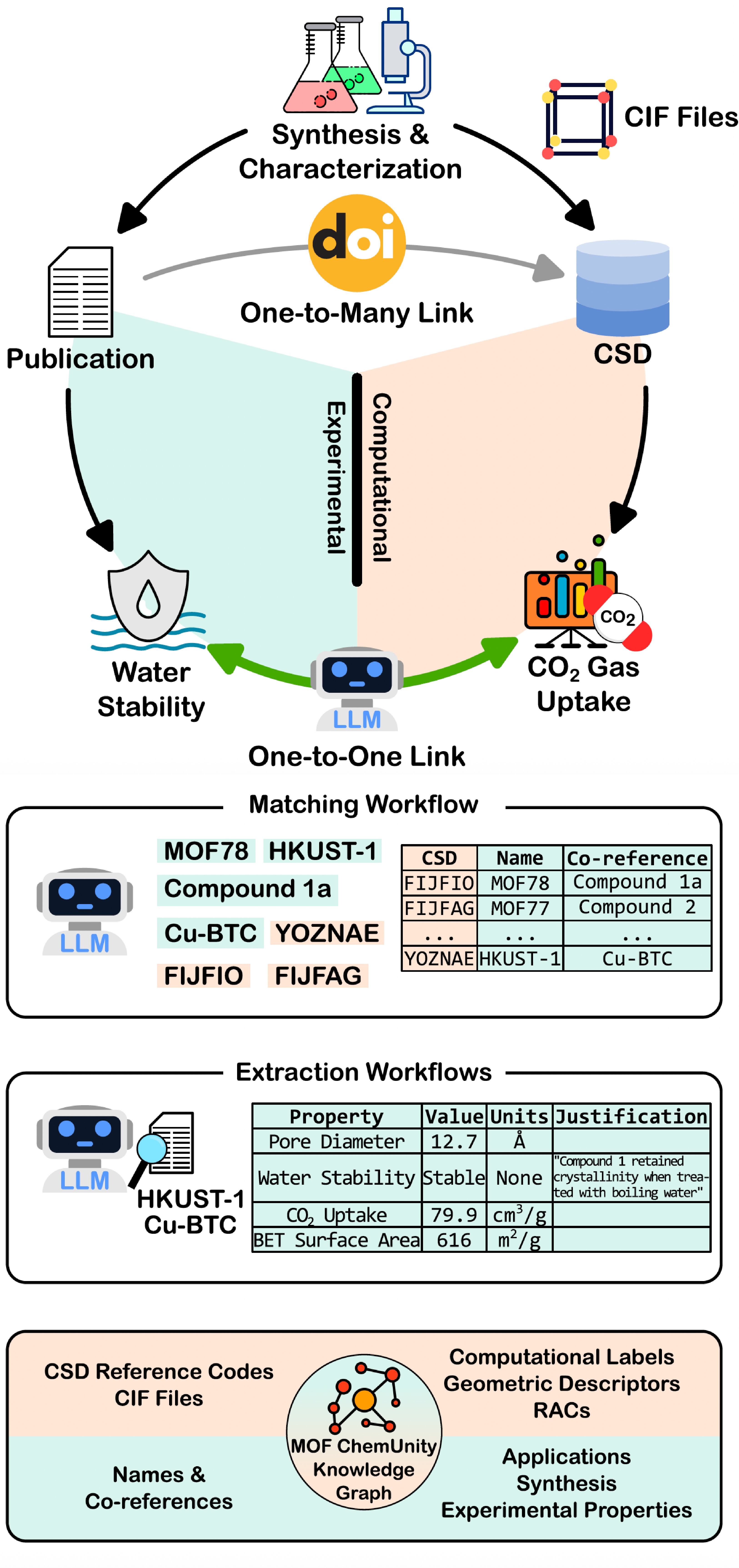} 
    \caption{The MOF-ChemUnity workflow.
LLM are used to link publications and CSD entries by extracting experimental properties and matching compound names across literature and structure files.
This structured data populates the knowledge graph which combines synthesis, applications and so on.
Reproduced from ref.~\citenum{pruyn_mof-chemunity_2025}, licensed under CC BY-NC 4.0.}
    \label{fig:mofchemunity}
\end{figure}

The discovery of advanced materials, such as metal-organic frameworks (MOFs)\cite{kitagawa_metalorganic_2014,furukawa_chemistry_2013,noauthor_nobel_nodate}, have become pivotal to breakthroughs in areas like energy storage, catalysis, and chemical separation.\cite{xing_highentropy_2025, li_advances_2024} However, the traditional labour-intensive and trial-and-error research paradigm in materials research moves at a slow pace.
Decades of research have produced a vast, yet fragmented, archive of knowledge scattered across millions of scientific literature, among other data repositories.\cite{stracke_rise_2024} Therefore, systematically extracting unstructured data with minimal manual effort is crucial for data collection and analysis.
Traditional techniques such as regular expressions (RegEx), part-of-speech tagging, and even transformer-based extraction methods rely heavily on rule-based structures or predefined structures, which makes them very rigid as they struggle to accommodate the full diversity of natural language expressions.\cite{bae_text_2025} In contrast, by leveraging their pre-trained knowledge, LLMs offer an advanced opportunity to understand the intricate chemistry language and textual context, enabling them to process unstructured data with higher flexibility.\cite{schilling-wilhelmi_text_2025,zhang_discovery_2025}

Building upon clean, actionable data as a foundation unlocks new possibilities in materials science research.
One of the ambitious goals in materials science research is to establish the structure-property relationships that govern material performance.
By training on vast and comprehensive datasets of chemical information, LLMs can potentially learn these intricate connections and provide valuable insights into the fundamental principles.\cite{noauthor_mllm_nodate,wang_survey_2025} Additionally, LLMs are undergoing a path from being a passive assistant toward becoming an active participant in the research process.
The most advanced applications now integrate LLMs as a central ``brain'' into research workflows, where these agentic systems can plan multi-step procedures, interface with computational simulation tools, and even operate robotic platforms.\cite{mitchener_kosmos_2025,darvish_organa_2025,amendible-barreto_dynamate_2025}  

It is inevitable that LLMs are revolutionising the materials science landscape.
This potential, however, has largely been explored using closed-source, commercial models such as OpenAI's GPT series.
Researchers often default to these models as they are production-ready, easy to use, and offer top-tier performance.
While these models are industry-leading, their closed-source nature presents many drawbacks: high costs for large-scale or high-throughput tasks, data privacy concerns, reduced reproducibility, and limited flexibility for model customisation.\cite{kumar_navigating_2025}.
In parallel, the open-source LLM ecosystem has expanded significantly. The release of Meta’s Llama 3 family\cite{grattafiori_llama_2024} in early 2024 marked the first time open-source models achieved true commercial-grade competitiveness with their closed-source ones.
This milestone established a strong foundation for both research and industry applications.
Subsequently, Alibaba’s Qwen\cite{yang_qwen3_2025} and Zhipu AI’s GLM series\cite{team_glm-45_2025} have made substantial progress toward matching and even surpassing proprietary models.
In this review, we outline the use of LLMs in materials science applications, with particular focus on MOFs.
We also examine the capabilities of rapidly emerging open-source models across these diverse tasks.
\section{Intelligent Data Extraction and Curation}
The traditional approach to discovering new materials has largely been driven by trial and error, which requires extensive experimentation and validation.\cite{thornton_materials_2017} With the efforts of thousands of dedicated researchers, a vast amount of valuable information has been accumulated over time.
However, they remain sporadic and scattered across different sources, therefore transforming this fragmented knowledge into a unified, standardised database can significantly facilitate the entire research process by enabling faster information retrieval and data-driven analysis.
In the work by Ghosh et al., an LLM-driven workflow was developed to autonomously draw out key thermoelectric properties (e.g., Seebeck coefficient. thermal conductivity) and associated structural properties, such as crystal class, space group, and doping strategy) from approximately 10,000 material science articles.\cite{ghosh_automated_2025} They also benchmarked different Gemini and GPT models and found GPT-4.1 mini offered the best cost-performance balance.
This effort resulted in the creation of the largest LLM-curated thermoelectric dataset which contains 27,822 temperature-resolved property records for a diverse class of materials.
A key strength of this work lies in its explicit focus of tables and their associated captions as distinct, high-value data sources, rather than relying solely on unstructured text.
Similarly, Li et al. developed an extraction workflow called ``ReactionSeek'' which is capable of directly interpreting reaction scheme images using a multimodal LLM (GLM-4V) and achieved an accuracy of 91.5\% when tested on a set of 42 diverse images.\cite{li_reactionseek_2025} These multimodal expansions effectively broaden the scope of accessible data and enable a more comprehensive understanding of scientific information.
It is worth noting that the model employed here are open-source and demonstrate impressive performance, highlighting the growing potential of community-driven models in specialised scientific applications.

\begin{figure}[htbp] 
    \centering
    \includegraphics[width=0.4\textwidth]{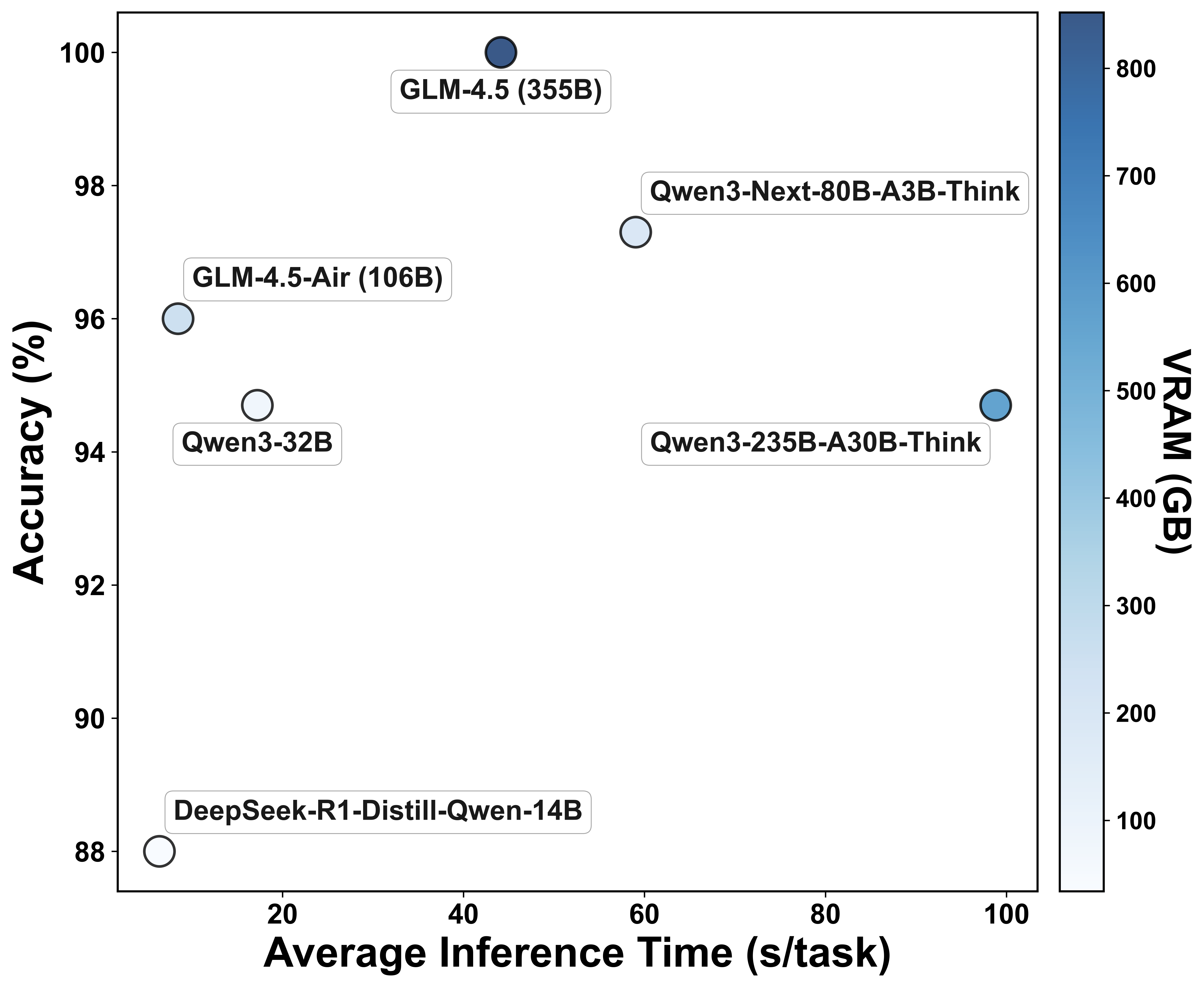} 
    \caption{Performance benchmark of open-source LLMs on the MOF-ChemUnity synthesis conditions extraction task.
Model performance is plotted across three key dimensions: accuracy (\%), average inference time (s/task), and estimated VRAM (GB) usage under bfloat16 precision (color scale).
Note that Qwen3 models exhibit significantly higher average inference times compared to similarly sized models.
This is attributed to their reasoning process prior to output generation.
All models were evaluated with thinking mode enabled.}
    \label{fig:result1}
\end{figure}

Towards the development of MOFs, Pruyn et al.
developed ``MOF-ChemUnity'' (Figure~\ref{fig:mofchemunity}) that not only extract key information such as material properties and synthesis procedures, but also links the various names used for these materials to their corresponding co-reference names and crystal structures.\cite{pruyn_mof-chemunity_2025} This linkage bridges between textual synthesis and property knowledge and the atomic-level structural insights.
Finally, the mined datasets form a knowledge graph that serves as a structured, scalable, and queryable foundation for materials discovery.
However, while this approach captures the important details such as the overall synthesis duration, it only extracts static attributes and ignore the sequential order and relationships among synthesis actions.
The work by Zhao et al. directly targets this gap by presenting an ``sequence-aware'' extraction, capturing the step-by-step experimental workflow as a directed graph, where each node represents an action (e.g., ``mix'', ``heat'', ``filter''), and edges define the experimental sequence.\cite{zhao_expert-guided_2025} This workflow achieved high F1-scores for both entity (0.96) and relation (0.94) extraction.
All of these studies highlight a significant shift from simple data extraction toward creating dynamic, AI-ready knowledge bases that enable sophisticated, data-driven discovery.
To demonstrate the performance of open-source models on these data extraction tasks, we reproduced the benchmark for six synthesis conditions provided by the MOF-ChemUnity code repository.\cite{noauthor_ai4chemsmof_chemunity_2025} The models tested included the Qwen3 and GLM-4.5 series, featuring both dense and Mixture-of-Experts architectures with sizes ranging from 14B to 355B parameters.
As shown in Figure~\ref{fig:result1}, most models achieved accuracies exceeding 90\%, with the largest model reaching 100\%.
This highlights the strong potential of open-source models which demonstrates their capability to effectively handle data-mining tasks.
Notably, small models such as Qwen3-32B yielded an accuracy of 94.7\%, suggesting that compact models can also handle the task effectively.
This is significant as these smaller models require far fewer computational resources;
Qwen3-32B, for instance, can be readily deployed on a standard Mac Studio with an M2 Ultra or M3 Max chip.
The original study divided full-text literature into smaller chunks during pre-processing to identify relevant experimental paragraphs, which were then fed to GPT-4o for extraction.
While this approach helps narrow the search space, it inevitably results in some loss of contextual detail, potentially omitting valuable features.
In contrast, we processed entire papers using open-source models, which enabled the capture of additional information dispersed throughout the document that may have been missed by chunk-based approaches.
\section{Predictive Power of LLMs}
\begin{figure}[htbp] 
    \centering
    \includegraphics[width=0.8\textwidth]{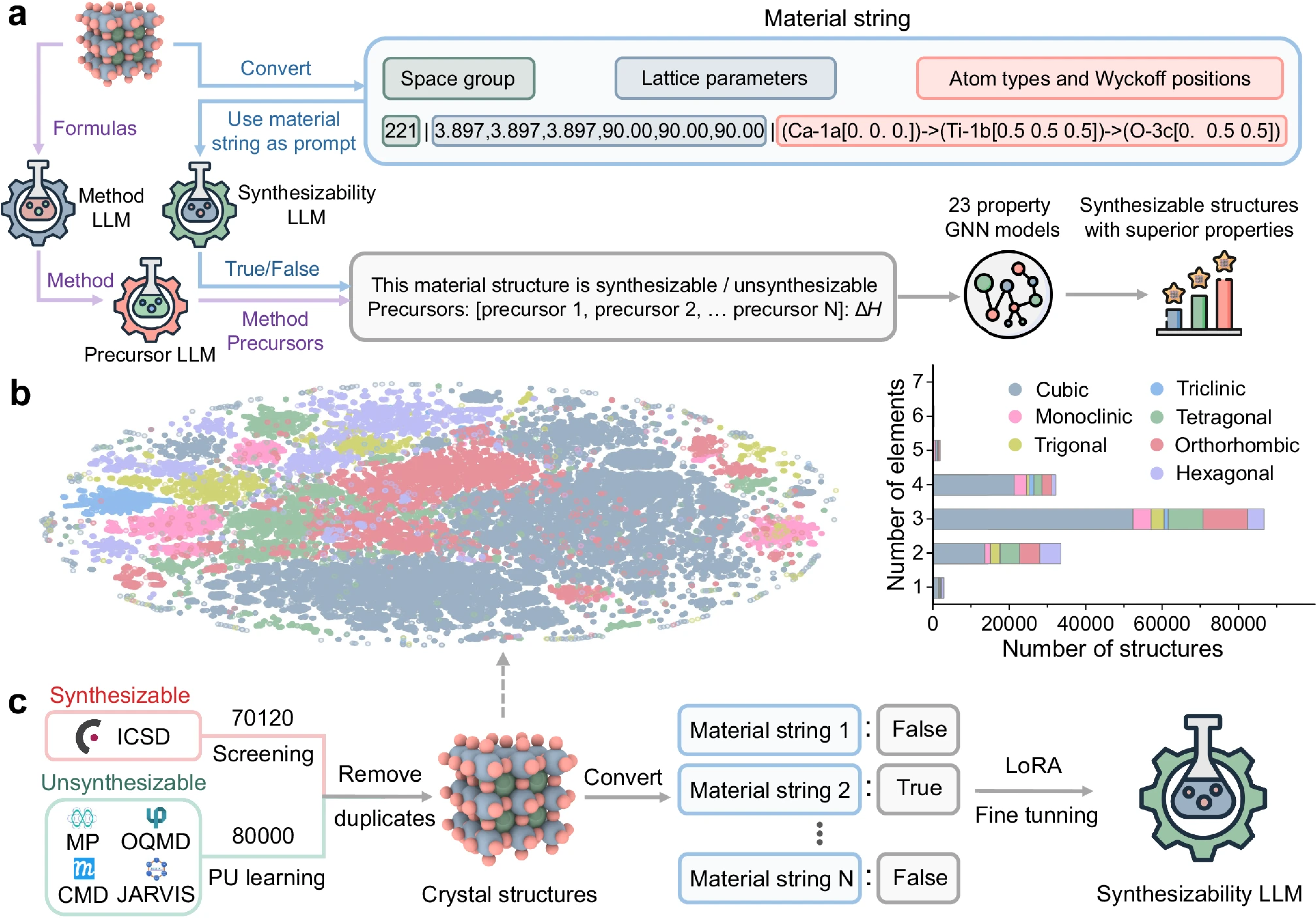} 
    \caption{Framework for predicting material synthesisability and synthesis routes.
(a) Material string encoding structural data is used to train a (c) "Synthesizability LLM".
Reproduced from ref.~\citenum{song_accurate_2025}, licensed under CC BY-NC-ND 4.0.}
    \label{fig:materialstring}
\end{figure}

Beyond simply reciting information Kang and coworkers developed a system called ``L2M3'' that not only extracts MOF synthesis conditions for database construction, but also provides a ``recommender'' tool by predicting synthesis conditions based on provided precursors by users.\cite{kang_harnessing_2025} As most LLMs are designed for general-purpose applications, they may not inherently excel at domain-specific tasks especially in material science where there is complex terminology or unique patterns.
To address this, they finetuned GPT-3.5-turbo and GPT-4o using only the textual formulas of MOF precursors and both models achieved a moderate similarity score of 82\% compared to true experimental conditions.
Further, the work by Liu et al. demonstrated the predictive performance for MOF properties by providing compositions as well as high-level structural features like node connectivity and topology through rich, natural language descriptions.\cite{liu_leveraging_2025} The finetuned model achieved 94.8\% accuracy in predicting hydrogen storage performance which is a substantial 46.7\% improvement over models using only the precursor names.
To better encode comprehensive atomic-level details for LLMs, Song and colleagues proposed a new material representation format called ``Material String'' (see Figure~\ref{fig:materialstring}), which is designed to be significantly shorter and more information-dense than standard crystal structure files like CIF or POSCAR.\cite{song_accurate_2025} This atomic-level representation encodes essential structure details such as space group, lattice parameters, and Wyckoff positions, allowing the complete mathematical reconstruction of a material’s primitive cell in 3D.
The finetuned model showed remarkable accuracy on the synthesisability test (98.6\%).
More importantly, it exhibited excellent generalisation, maintaining an average accuracy of 97.8\% even when tested on complex experimental structures with up to 275 atoms and is far beyond the 40-atom limit of its training data.
The model also achieved impressive performance on prediction of synthesis routes (91.0\%).
All of these results collectively underscore the ability of LLMs to learn and capture complex structural patterns or property features.

\begin{figure}[htbp] 
    \centering
    \includegraphics[width=0.4\textwidth]{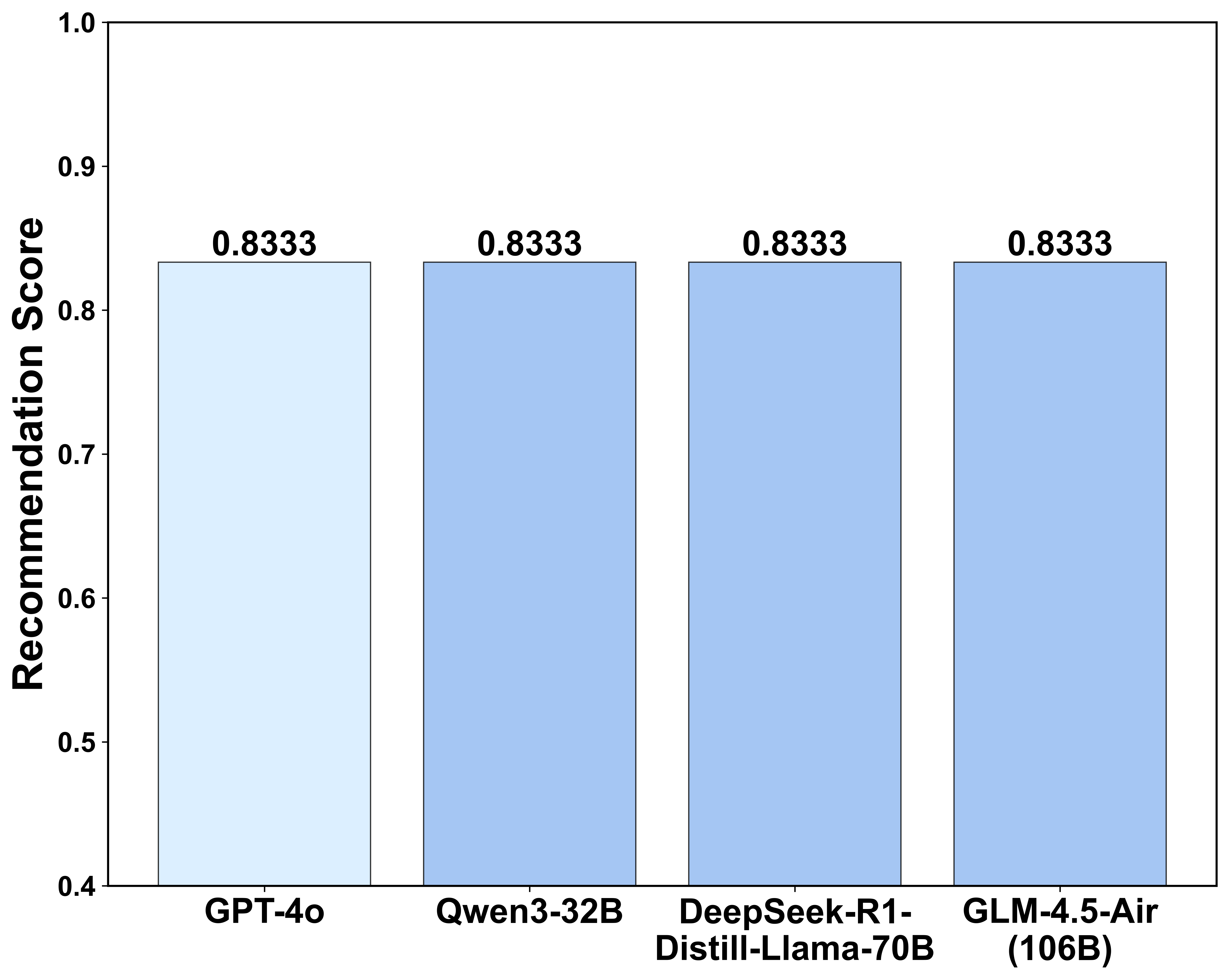} 
    \caption{Comparison of synthesis condition recommendation scores (median per-sample) for different finetuned open-source models.
The result for GPT-4o (left) is the reported score from the original study.}
    \label{fig:result2}
\end{figure}

To evaluate the capability of open-source models on prediction tasks, we fine-tuned three models of varying model sizes and architectures on the training dataset provided by L2M3.\cite{noauthor_taeun8991l2m3_2025}.
As the official test set was unavailable, we split further split the training dataset into 85\% (4,990 samples) and 15\% (1,039 samples) for training and evaluation, respectively.
We employed Low-Rank Adaptation (LoRA) with a rank of 32 for efficient fine-tuning, which enabled the largest model tested, GLM-4.5-Air, to fit within four AMD Instinct MI250X Accelerators.
When combined with 4-bit quantisation, the finetuning could be accommodated on only two MI250X with a minor loss in accuracy.\cite{dettmers_qlora_2023} All models achieved a median score identical to that reported for GPT-4o (Figure~\ref{fig:result2}).
This result further demonstrates that open-source models can effectively learn underlying data patterns at a level comparable to closed-source models.

Upon closer inspection of our results we found the dataset is highly imbalanced (Supporting Information), which makes it unsuitable for reliable model training.
This imbalance can bias the model toward the majority class, producing deceptively high accuracy while failing to learn meaningful patterns in the minority class.
We also observed that one component of the similarity metric simply repeats the input precursors, yielding nearly 100 percent accuracy and further inflating the median score.
These findings suggest that the practical usability of models trained solely on non-structural precursor formulas or names is very limited, which reinforce the importance of incorporating high-quality structural representations as input features.
This analysis also underscores the critical need for transparent reporting and data sharing.
We observed that most studies in this domain lack sufficient detail for experimental reproduction, making it challenging to validate results or build upon existing work.
\section{LLM Agents for Chemical Discovery}
\begin{figure}[htbp] 
    \centering
    \includegraphics[width=0.8\textwidth]{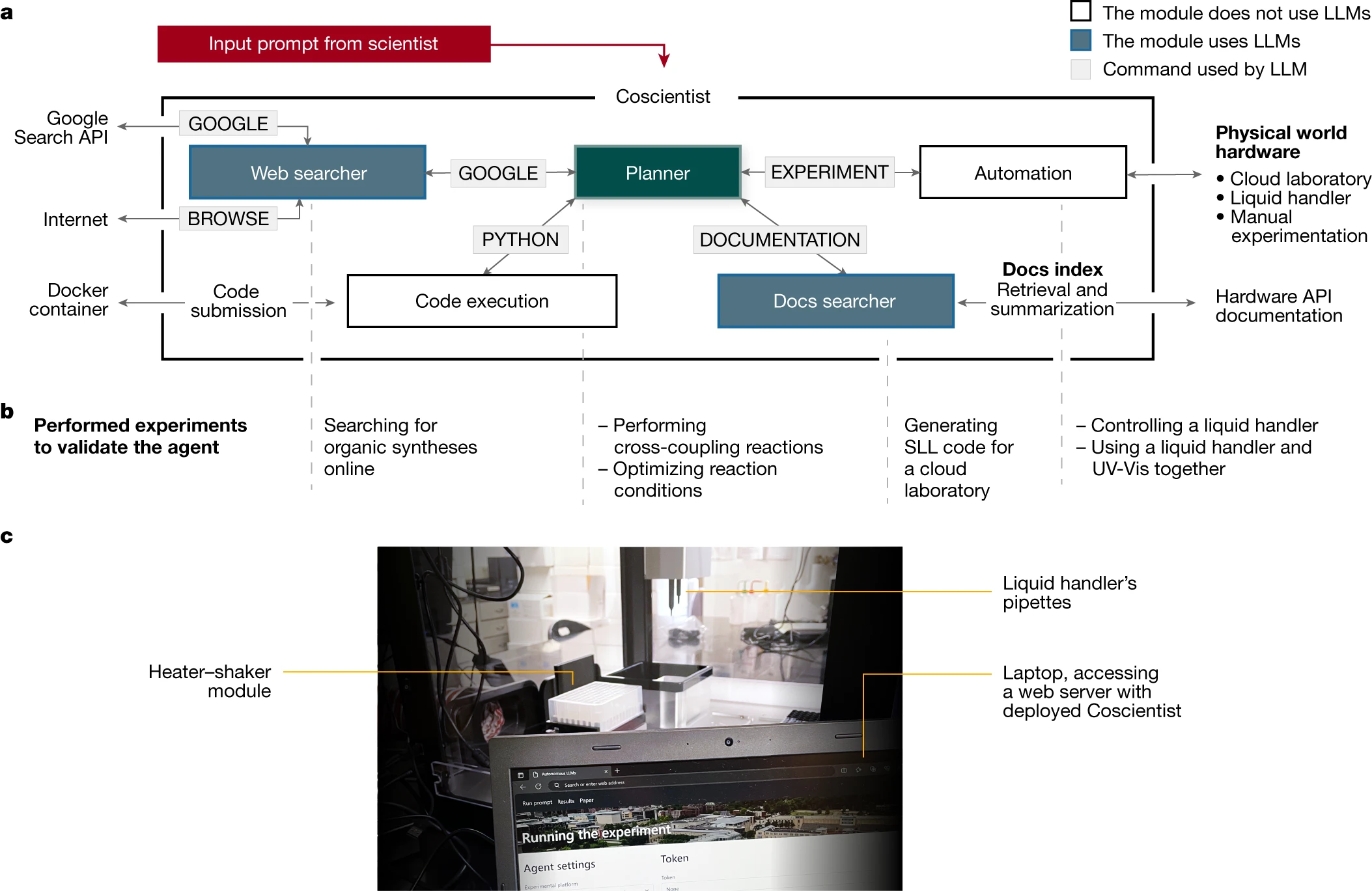} 
    \caption{A comprehensive system featuring a central LLM-based ``Planner'' that orchestrates and manages the entire research workflow.
Reproduced from ref.~\citenum{boiko_autonomous_2023}, licensed under CC BY 4.0.}
    \label{fig:coscientist}
\end{figure}

The development of LLMs is increasingly transitioning from single-model assistance towards sophisticated, multi-agent systems seemingly capable of performing complex human tasks.\cite{wang_survey_2024} This transformation is envisioned to revolutionise materials science, where most material research has long been constrained by laborious cycles of hypothesis formulation, pre-experimental research, and experimentation.
By integrating LLMs into the research system, we can minimise unnecessary human intervention and accelerate the pace of discovery.
One of the most impactful roles of LLMs is supporting researchers in the exploration and refinement of their research ideas.
As demonstrated by the SciAgents framework, LLMs can navigate vast knowledge base to uncover previously unseen connections between disparate scientific concepts, which leads to the generation of novel hypotheses.\cite{ghafarollahi_sciagents_2025} In addition, this framework can iteratively refine and elaborate on initial concepts.
Different AI agents within the  SciAgents framework can expand upon a hypothesis by adding quantitative details, suggesting specific modelling or simulation priorities, and providing comprehensive critiques that identify strengths, weaknesses, and areas for improvement.
This feedback loop effectively mimics and accelerates the traditional scientific process of discussion and peer review, ensuring that the resulting ideas are not only innovative but also scientifically rigorous.
LLMs also hold great potential to function as central coordinators, effectively connecting researchers with complex computational tools before the experimental stage.
By leveraging their advanced reasoning abilities, LLM-based agents can understand and process queries in natural language, eliminating the need for rigid, formal syntax or other technical knowledge for using a computational tool.
The ChatMOF system exemplifies this by orchestrating a sophisticated pipeline built on three core components: an agent, a toolkit, and an evaluator.\cite{kang_chatmof_2024} The toolkit contains a combination of the recognised databases such as QMOF and CoREMOF, and also a machine learning model that predicts material properties (e.g., hydrogen diffusivity), and a genetic algorithm tool for generating new combination of materials.
When a user submits a query, the agent (powered by an LLM like GPT-4) functions as the ``brain''.
It analyses the request, formulates a multi-step plan to solve the problem, and selects the appropriate instrument from its toolkit.
The evaluator then assesses the output from the tool and synthesises it into a final, coherent answer for the user.
Moreover, coupling LLM agents with laboratory automation and robotic synthesis platforms can close the loop between computation and experiment.
In the work by Boiko et al., an AI system named Coscientist was developed to autonomously design, plan, and perform complex experiments.\cite{boiko_autonomous_2023} Driven by GPT-4, the system functions as a central agent that coordinates a suite of tools for internet and documentation search, code execution, and experimental automation (Figure \ref{fig:coscientist}).
Similarly, Song et al. introduced ChemAgents, a multiagent system powered by an open-source model Llama-3.1-70B.\cite{song_multiagent-driven_2025} This system also features a central ``Task Manager'' that coordinates four highly specialised agents: Literature Reader, Experiment Designer, Computation Performer, and Robot Operator.
Each agent is explicitly linked to a foundational resource, such as a literature database or an automated lab, which makes it a highly flexible framework that is able to execute tasks ranging from literature review to robotic operation.
Both Coscientist and ChemAgents illustrate the view of leveraging LLM agents for general-purpose, ``horizontal'' platforms that can adapt to automate a wide range of detailed tasks.

Continuing the development of multi-agent systems, their scope has been further refined to focus on material discovery and optimisation.
Zheng et al. developed a ChatGPT research group where seven distinct LLM-based assistants collaborate with a single human researcher, who only needs to specify the research objective through prompts.\cite{zheng_chatgpt_2023-1} As a result, this system successfully accelerated the finding of optimal synthesis conditions for MOFs and COFs by coupling AI agents with Bayesian optimisation, balancing the exploration and exploitation of a vast parameter space and reducing millions of potential conditions to a manageable number.
In a related effort focused on de novo discovery, MOFGen was presented as a system of agentic AI dedicated to discovering novel, synthesisable MOFs.\cite{inizan_system_2025} This system employs a pipeline of specialised agents such as an LLM called LinkerGen proposes novel compositions, a diffusion model generates 3D crystal structures, and other agents that perform quantum mechanical filtering and synthesisability analysis.
This generative approach led to the successful synthesis of five ``AI-dreamt'' MOFs.
These ``vertical'' applications demonstrate how such agentic structures can be specialised to solve deeper problems within a single domain.
Moreover, all four systems highlight a core design principle that their strength lies not in the LLM alone, but in its ability to plan, delegate, and distribute sub-tasks across the entire research environment.
While the emergence of agentic systems represents a fascinating step toward the future of autonomous research, most current LLM-based agents in this domain still rely heavily on closed-source models, which introduces some trade-offs.
Commercial models can cause significant financial cost and accessing them through APIs can expose systems to instability and security risks.
APIs are prone to outages, disruptions, and unexpected updates that may compromise reliability, and their dependence on internet connectivity makes it difficult to ensure data privacy and compliance with confidentiality protocols, which is a major concern for sensitive experiments or proprietary research data.
In contrast, the rapid advancement of open-source LLMs is beginning to change this landscape.
Models such as GLM-4.5 and Qwen3-235B-Thinking-2507 already demonstrate agentic and reasoning performance comparable to that of leading closed-source counterparts.\cite{team_glm-45_2025} Therefore, selecting an appropriate system should strike a balance between performance, resource availability, and operational independence, and open-source development may ultimately pave the way for sustainable, locally deployable autonomous research systems.
\section*{Conclusions and Outlook}
In conclusion, the integration of LLMs across data extraction, predictive modelling, and agentic system demonstrates an emerging capability to interpret complex chemistry text, learn structure–property relationships, and orchestrate research workflows.
The studies reviewed here together with our benchmarking results confirm that open-source models can achieve comparable performance to closed-source systems while offering superior transparency, flexibility, and reproducibility.
Collectively, these advances mark a transition from isolated, task-specific LLM applications toward unified, AI-driven research ecosystems that accelerate the deconstruction of complex scientific knowledge, minimise repetitive manual effort, and open new frontiers for exploration and discovery.
However, evaluating the reliability of highly autonomous systems remains a challenge.
While recent studies have attempted to benchmark the agentic abilities of LLMs such as tool calling,\cite{noauthor_chemagent_nodate} there is still not yet a comprehensive framework for assessing performance beyond one-step reasoning accuracy.
Complex agentic systems require the integration of planning, execution, and adaptive decision-making, which current benchmarks do not capture.
Moreover, existing metrics predominantly measure procedural correctness rather than the holistic resilience of reasoning when confronted with uncertainty or failure.
Developing new benchmarks is therefore crucial to clarify the true competency of models acting as ``researchers''.
Such frameworks would not only guide the need for domain-specific fine-tuning, but also be essential for building trustworthy autonomous systems that can be confidently adopted in real-world scientific research.
\section*{Conflicts of interest}
There are no conflicts to declare.

\section*{Data availability}
Code and data related to benchmarking and fine-tuning tools are available on Zenodo: \href{https://doi.org/10.5281/zenodo.17548056}{10.5281/zenodo.17548056}.
\section*{Acknowledgements}
J.D.E is the recipient of an Australian Research Council Discovery Early Career Award (project number DE220100163) funded by the Australian Government.
Phoenix HPC service at the University of Adelaide is thanked for providing high-performance computing resources.
This research was supported by the Australian Government's National Collaborative Research Infrastructure Strategy (NCRIS), with access to computational resources provided by
Pawsey Supercomputing Research Centre through the National Computational Merit Allocation Scheme.
We thank Dr Fabien Voisin (Phoenix HPC, University of Adelaide) for his assistance managing our resource needs.


\end{document}